\documentclass[final]{cvpr}

\usepackage{float}
\usepackage{times}
\usepackage{epsfig}
\usepackage{graphicx}
\usepackage{amsmath}
\usepackage{amssymb}

\usepackage[pagebackref=true,breaklinks=true,colorlinks,bookmarks=false]{hyperref}

\def\cvprPaperID{****} 
\def\confYear{CVPR 2021}

\begin{document}

\title{se-Shweshwe Inspired Fashion Generation}

\author{Lindiwe Brigitte Malobola\\
University of the Witwatersrand\\
{\tt\small brigitte.malobola@gmail.com}
\and
Negar Rostamzadeh\\
Google Research\\
{\tt\small nrostamzadeh@google.com}
\and
Shakir Mohamed\\
DeepMind\\
{\tt\small shakir@deepmind.com}
}

\maketitle
\begin{abstract}
Fashion is one of the ways in which we show ourselves to the world. It is a reflection of our personal decisions and one of the ways in which people distinguish and represent themselves. 
In this paper, we focus on the fashion design process and expand computer vision for fashion beyond its current focus on western fashion. We discuss the history of Southern African se-Shweshwe fabric fashion, the collection of a se-Shweshwe dataset, and the application of sketch-to-design image generation for affordable fashion-design. The application to fashion raises both technical questions of training with small amounts of data, and also important questions for computer vision beyond fairness, in particular ethical considerations on creating and employing fashion datasets, and how computer vision supports cultural representation and might avoid algorithmic cultural appropriation.
 \end{abstract}
 
\section{Introduction}
Fashion has drawn a lot of attention from researchers in computer vision in recent years with a growing number of papers and workshops dedicated to this topic. There has been rapid development in fashion-related work ranging from fashion analysis, detection, fashion synthesis, and recommendation. This rapid development has led to a number of publicly available fashion datasets that are suitable for the development and application of machine learning~\cite{rostamzadeh2018fashion, liu2016deepfashion, zakizadeh2018improving, yang2014clothing, zheng2018modanet, loni2014fashion}. However, to the best of our knowledge, existing works on machine learning for fashion have not yet considered applications to African-inspired fashion and are limited to western forms of fashions. This limited diversity in existing data is linked to the general under-representation of non-western cultures and art forms, and contributes to the general harms that arise from perpetuating single stories about people and cultures.\\

In addressing this limitation our contributions in this work are: firstly, we curate the first dataset that considers African fashion, in particular, the creation of a fashion dataset representing different designs of Southern African modern Shweshwe fashion dresses; Secondly, we make initial contributions in the training of sketch-to-design GANs from small datasets, to explore the problem of generating African-inspired fashion from sketches; Thirdly, we explore important questions for computer vision beyond fairness, in particular questions of AI in relation to cultural representation and algorithmic appropriation.


\section{Towards Affordable Good Quality Cultural Fashion}

Fashion represents one of the long-lasting modes of preserving and celebrating culture and history. In parts of the world, fashion is used to signify different social groups and status. In different parts of the African continent people use fashion to distinguish the many different cultural traditions, and it is also used as a celebration of their unique histories, on special occasions, and on a daily basis for some. Our focus in this paper is on fashion designs based on one of the Southern African traditional wear.\\

African traditional wear has had a large influence on global fashion brands, although often not acknowledged nor given recognition of its rooted cultural significance. \\ 
We look at the fashion design process, which often begins with conceptual drawings. Fashion designers translate these concept drawings into a pattern, create a basic version of the garment with inexpensive cotton, and then finally samples are made with desired fabrics. This process is costly and can stifle productivity in the cultural fashion sector where, in a country like South Africa, 70\% of cultural fashion enterprises are informal~\cite{SEDA_2018} with an average annual turnover R96,439\footnote{Where R96,439 = US\$6,52 on average}~\cite{snowball2020creative}. This poses a significant challenge for the cultural fashion sector to take advantage of the growing market in cultural fashion for everyday use~\cite{snowball2020creative}.\\

In this work, we develop models that can be used as tools for generating African inspired fashion images from conceptual drawings. A sample generating system will help designers in informal markets communicate their concepts faster and secure orders from customers before creating a sample garment. This blend of product creativity and digital advancement can result in a strong competitive edge for the struggling South African textile industry as a whole.\\
 
\section{se-Shweshwe Fabric and its History}
se-Shweshwe\footnote{se-Shweshwe is Pronounced: ci-sh-wesh-where } is a dyed cotton fabric characterised by elaborate flower arrangements, square, circular, stripes or diamond geometric prints. The fabric was originally dyed indigo but is now being produced in different colours. se-Shweshwe is one of the fabrics that are central to Southern African traditional clothing. It is commonly worn at traditional ceremonies such as weddings and lobola ceremonies.  \\

Historically in African culture se-shweshwe has been used to make several cultural apparels. These apparel range from aprons, skirts and dresses among others.  The African cultures in question that mostly make use of the se-shweshwe are the Batswana\footnote{Batswana is Pronounced:Baat-swa-na}, AmaXhosa\footnote{Pronunciation: \url{https://www.youtube.com/watch?v=Trq_gIe1v04}} and the Basotho\footnote{Basotho is Pronounced:Baa-soo-to}. Se-shweshwe is mostly worn by newly married women known as uMakoti\footnote{uMakoti is Pronounced: oo-mark-oot-ee}. Further, in the Xhosa culture, the se-shweshwe has been embraced to be a part of ochre-coloured blanket clothing. Herero women in Namibia use se-Shweshwe to make some of their dresses \footnote{\url{shorturl.at/aTX45}}. In addition se-shweshwe is used in modern South African fashion clothing design for all genders from different cultures . 
\\

The se-Shweshwe fabric has a  long and rich history. The intellectual contribution can’t be attributed to a single individual, but results from, and defines, the cultural evolution of specific groups of people. Se-shweshwe’s earliest origin can be dated all the way back to the trend of colourful, floral Indian cotton called Indienne that spread very quickly in the mid-16th century through Europe \footnote{\url{shorturl.at/kmBGN}}. The cloth has been imported from Europe to South Africa.
The trademarked cloth is being manufactured today by Da Gama Textiles in the Zwelitsha township in the Eastern Cape since 1982 \footnote{\url{shorturl.at/hrBYZ}}. Da Gama textiles has bought rights to the most popular brand called Three Cats.\\  
Present-day se-Shweshwe is used as an input into the production of custom made garments and accessories on both an industrial scale and at a small-scale. In particular, the fabric is part of the production of numerous micro enterprises, which are often operating informally (unregistered businesses) producing custom-made garments for traditional ceremonies and festivities, such as weddings.

\begin{figure*}[]
    \begin{center}
    \makebox[\linewidth]{
    \includegraphics[height= 8cm]{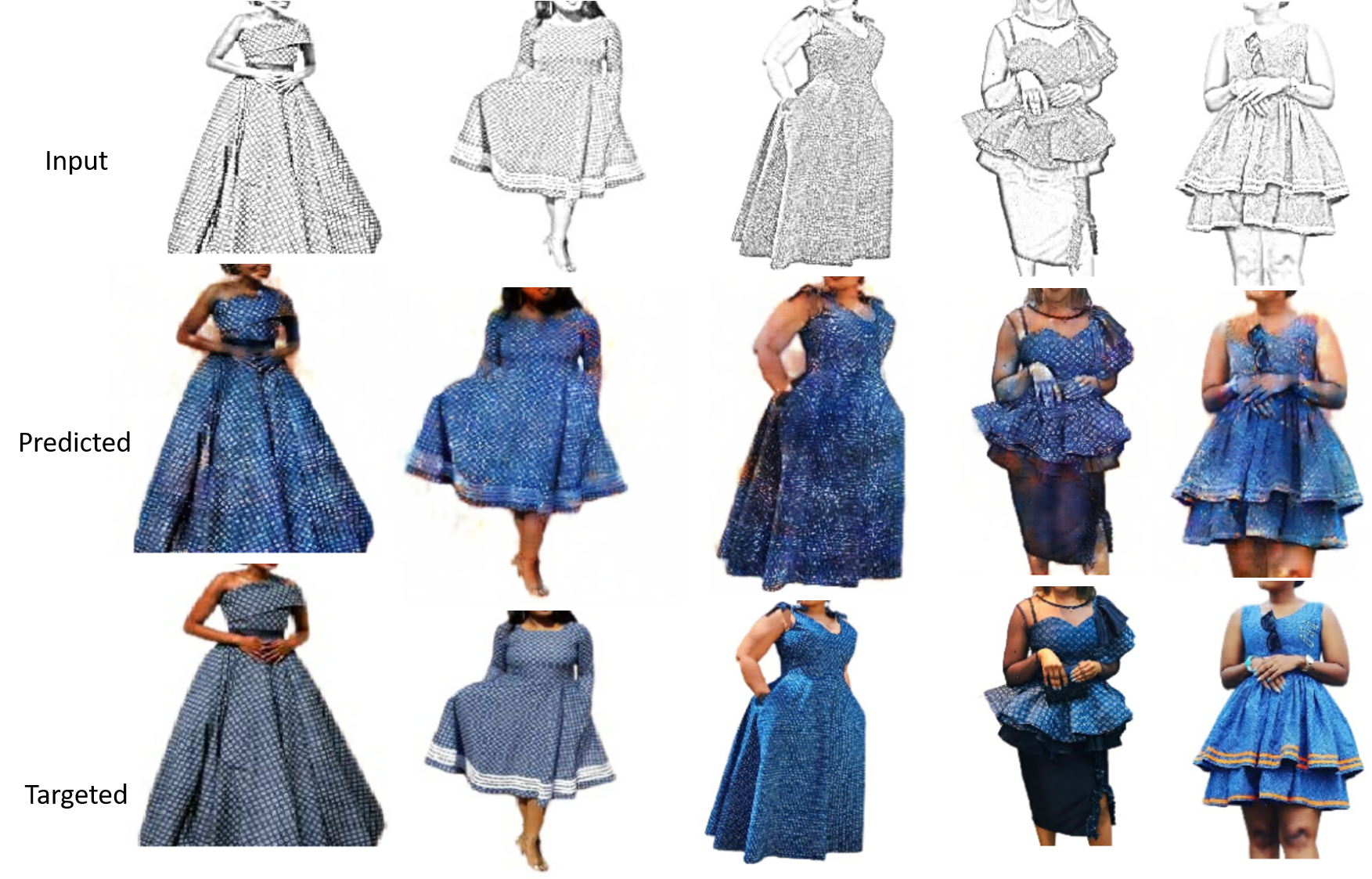}}\\
    \caption{Examples of paired images in the se-Shweshwe dataset and their respective prediction outputs. The predictions are outputs of the sketch to image translation modified Pix2pix\cite{isola2017image} model.}
    \label{expshw}
    \end{center}
\end{figure*}


.

\section{Se-shweshwe Dataset}
The dataset is made up of images of modern African fashion items and corresponding sketches split into train and test subsets. The fashion items, most of which are dresses, are  made of the popular se-Shweshwe fabric. We describe the collection of the dataset, how it was cleaned, and how we created a paired sketch-image data. \\

\paragraph{Data Collection and Cleaning}
We collected photographs of se-Shweshwe dresses using images obtained from an image search engine\footnote{Google Images \url{https://www.google.com/imghp}} using specific keywords. The query keywords used were `se-Shweshwe dresses', `Batswana traditional dresses' and `Basotho traditional dress',  with the resulting photos downloaded from social media websites, forums, blogs, and other content sources. A total of 640 photographs with noisy backgrounds were obtained from the initial images search.\\
 
To clean the dataset, we identified and removed duplicates, removed unusable images of low resolution or irrelevant objects. Images were all cropped to be of size $256 \times 256 \times 3$. Faces in the dataset were removed for privacy purposes. A total of 500 se-Shweshwe clothing images of varying designs were retained to make up the final se-Shweshwe dataset. 
Models trained on images with noisy background generated poor quality images. While reliance on background in classification tasks is still an open problem~\cite{xiao2020noise}, we have seen that a standardized background on images can significantly contribute to the quality of the generated images~\cite{rostamzadeh2018fashion}. This motivated the removal of the noisy background in our images. To remove background from images we made use of online background removing tools\footnote{\url{https://www.remove.bg/} and \url{https://removal.ai/}} with some human interaction for visual analysis and making manual corrections. 

\paragraph{Creating Sketches}
Our application will be the generation of fashion images from sketches. To create a dataset of paired sketches and dress images, we used an off-the-shelf method that converts images to pencil sketches\footnote{\url{https://www.youtube.com/watch?v=vS2ubdiAXvg&t=11s}}.  
At the end of this process we retained 500 sketches and 500 corresponding images. These two sets were divided into subsets of 400 elements for training  and 100 elements for testing in a paired manner.
 
\section{Related Work}

Variations of GANs are widely used in digital and generative arts, such as automatic anime character creation~\cite{jin2017towards}, digital image collage creation relying on StyleGAN~\cite{karras2019style} by Simon and Shiri\footnote{\url{http://www.aiartonline.com/highlights-2020/joel-simon-tal-shiri/}}, and paintings to photographs translation~\cite{zhu2017unpaired}.

Pix2pix~\cite{isola2017image} is a conditional GAN, that has been adapted for multiple artistic work due to its success in various paired image translation tasks. These range from colourising black-and-white images, transforming daytime scenes into night, edge to image translation and many more.\\

Generative models have also been used as a tool in fashion and style generation ideas, and designs. There are multiple fashion datasets that are created either in generative fashion or in fashion image retrieval. Fashion-Gen\cite{rostamzadeh2018fashion} is a dataset on text to image translation, helpful for generating fashion designs from descriptions. Fashion-IQ~\cite{guo2019fashion} is a dataset for natural language based image retrieval systems. DeepFashion~\cite{liu2016deepfashion} is a large scale dataset with fine-grained annotation helpful for a variety of tasks related to fashion. Geostyle~\cite{mall2019geostyle} is a fashion dataset that was used to try to model fashion migration and influence in some parts of the world~\cite{al2020paris}.
In all these fashion datasets, the origins of the fashion items are not given, and they mainly consist of the western clothing styles.  

\section{Pix2pix Sketch to Image Translation}
Given the very small size of our data, to improve our model performance we increased the variety of the training data by resizing the training images to bigger height and width of size $286 \times 286$ then randomly cropped them to the target size of $256 \times 256$. We also applied random horizontal flipping, random $\pm 15$ degree rotations and added salt-pepper noise.\\ 
Since we focus on an affordable way to create a fashion design image from its sketch, we adopt the Pix2Pix paired image translation work, and make modifications to its original description to make it suitable for our purpose.
We used the hinge loss function instead of the binary cross-entropy loss. 
Instead of a constant learning rate used in the original paper we use a cyclical learning rate for both the Discriminator and the Generator. The cyclical learning rate far out performs the best constant learning rates considered in this study.
 
\begin{figure}[]
    \begin{center}
    \includegraphics[width = 4.1cm]{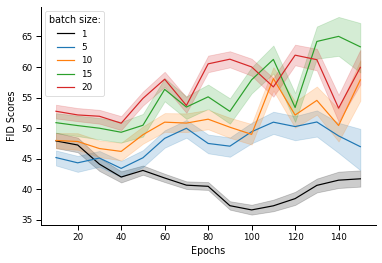}
    \includegraphics[width = 4.1cm]{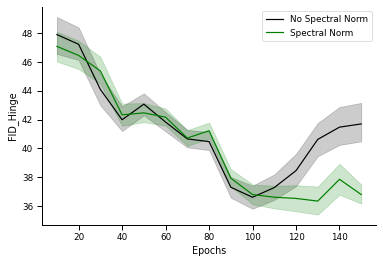}
    \caption{(Left)Increasing batch sizes. (Right)Spectral Normalization. The line graphs represent average FID scores of 10 runs and the shaded area represents the 99\% confidence intervals.}
    \label{batchsize}
    \end{center}
\end{figure}

We considered several batch sizes, but used a batch size of 1. Due to the small dataset we were training with, increasing the batch size for the baseline model greatly harmed the model performance, possibly due to overfitting. Fig. \ref{batchsize} shows that simply increasing the batch sizes by a factor of 5 results in higher FID scores at all the stages of training that were tested and the poor performance for higher batch sizes. \\
    
\begin{figure}[]
   \centering
   \includegraphics[width = 4.1cm]{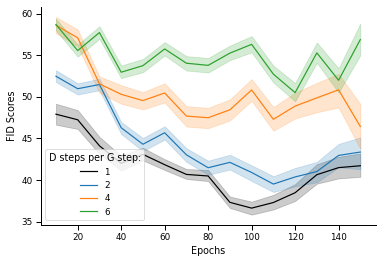}
    \includegraphics[width = 4.1cm]{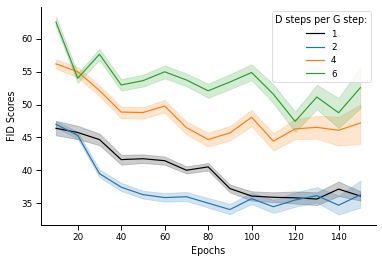}
     \caption{(Left)Number of \textbf{D} steps per \textbf{G} step experiment, first without spectral normalization. (Right)Number of \textbf{D} steps per \textbf{G} step experiment with spectral normalization.}
    \label{b1b1}
\end{figure}

The weights of the generator were normalized using spectral normalization~\cite{miyato2018spectral}. Spectral normalization in combination with 2 discriminator (\textbf{D}) steps works well. See Fig. \ref{b1b1}. We did an ablation study to analyse the effect of spectral normalization and 2,4 \& 6 \textbf{D} steps   independently and found that they don't help. See Fig. \ref{b1b1} and Fig. \ref{batchsize}.
The discriminator was given a higher learning rate interval than the Generator and took two discriminator steps per generator step.
    
\section{Ethical considerations in generative arts and the use of Se-shweshwe Dataset}

Generative models could have a positive impact on society by helping with cultural and identity representations. Aljowaysir, in Salaf\footnote{http://www.aiartonline.com/highlights-2020/nouf-aljowaysir/}, uses StyleGAN to reveal the political and social aspects of her past generations that is missed from the collective memory. Jake Elwes, by ``Zizi and Me"\footnote{https://www.jakeelwes.com/project-zizi-and-me.html}, relying on DeepFake, addresses challenges and discriminations against Queers and drag artists. se-Shweshwe inspired fashion generation, also represents a significant part of Southern African cultures, to broader communities. However, using generative models on representing cultures, adds a layer of complexity to a well-known potential problem, known as cultural appropriation. 

Cultural appropriation is when traditional knowledge and culture are adapted out of context, or re-purposed by members of another culture or identity~\cite{young2010cultural, unit1996making}. The harms can be amplified, when members of a dominant culture appropriate from disadvantaged minority cultures~\cite{young2009nothing}.\\
What makes cultural appropriation problematic in most cases is that cultural elements are often presented in a manner that strips off their cultural significance and meaning, and this can be specifically amplified using automated algorithms such as generative models.\\
``Textile craftsmanship is part of cultural heritage and has been an important element in building cultural identities. This is reflected in the traditional garments of different communities and indigenous people worldwide.''-\footnote{\url{https://www.culturalintellectualproperty.com/cultural-sustainability-in-fashion}}.\\
With this understanding that cultural textile craftsmanship is very closely tied with cultural identity, misrepresentation of cultural elements harms the sustainability of that culture and thus robbing future generations of the capacity of understanding and living the meanings and values of their heritage\footnote{\url{https://www.youtube.com/watch?v=twHCsVPupXo}, Monica Boța-Moisin, founder of Cultural 
Intelectual Property Rights initiative, which influenced a lot if our thinking about the concept of cultural sustainability in generative fashion. More of her work can be found here: \url{https://www.culturalintellectualproperty.com/library}}.

Working with se-Shweshwe fashion generation, brings a significant and important question. Where does the boundary of \textit{algorithmic cultural appropriation} and appreciation or representation sit?

How can an artist draw an inspiration and depict a cultural other? In words of Helen Fang\footnote{\url{https://www.youtube.com/watch?v=4wY5S3pfPis}}, an artist can still tell a story through informed ways that do not limit creativity. She also adds that appreciation and representation have to come from a respectful intent, with a thorough consideration of impact and awareness of contextual complexities such as commercialization, power dynamic and cultural sustainability.\\

Given the rate at which algorithms can quickly generate designs when compared to non-digital ways of creating designs and the scale at which this can be achieved, algorithmic fashion designers are in a position of power and advantage. This power and privilege comes with great responsibility.

Besides the significant problem of cultural appropriation, when the training data consists of a massive number of images created by other artists, credit assignment is a crucial question. Where do we draw the line between memorization (copying an art piece) and sourcing inspiration? How should the credits be attributed?



\bibliographystyle{unsrt}
\bibliography{bib}

\end{document}